\documentclass[sigplan,screen,authorversion,nonacm]{acmart}

\usepackage[a-1b]{pdfx}

\usepackage{dsfont}

\usepackage{fancyhdr}
\AtBeginDocument{%
    \addtolength{\footskip}{2.0\baselineskip}%
    \fancyfoot[L]{\textit{\textbf{Preprint --- do not distribute.}}}%
}

\AtBeginDocument{%
  \providecommand\BibTeX{{%
    \normalfont B\kern-0.5em{\scshape i\kern-0.25em b}\kern-0.8em\TeX}}}

\setcopyright{acmcopyright}
\copyrightyear{2020}
\acmYear{2020}
\acmDOI{10.1145/1122445.1122456}

\acmConference[FATE/MM '20]{FATE/MM '20: ACM MM 2020}{October, 2020}{US}
\acmBooktitle{FATE/MM ’20, October, 2020, US}
\acmPrice{15.00}
\acmISBN{978-1-4503-XXXX-X/18/06}

\begin{document}

\title{Balancing Fairness and Accuracy in Sentiment Detection using Multiple Black Box Models}


\author{Abdulaziz A. Almuzaini}
\affiliation{%
  \institution{Department of Computer Science}
  \streetaddress{}
  \city{Rutgers University}
  \country{New Brunswick, NJ}
  }
\email{aaa395@rutgers.edu}

\author{Vivek K. Singh}
\affiliation{%
  \institution{School of Communication and Information}
  \streetaddress{}
  \city{Rutgers University}
  \country{New Brunswick, NJ}
  }
\email{v.singh@rutgers.edu}

\renewcommand{\shortauthors}{Almuzaini and Singh.}

\begin{abstract}
Sentiment detection is an important building block for multiple information retrieval tasks such as product recommendation, cyberbullying detection, and misinformation detection. Unsurprisingly, multiple commercial APIs, each with different levels of accuracy and fairness, are now available for sentiment detection. While combining inputs from multiple modalities or black-box models for increasing accuracy is commonly studied in multimedia computing literature, there has been little work on combining different modalities for increasing \textit{fairness} of the resulting decision. In this work, we audit multiple commercial sentiment detection APIs for the gender bias in two-actor news headlines settings and report on the level of bias observed. Next, we propose a "Flexible Fair Regression" approach, which ensures satisfactory accuracy and fairness by jointly learning from multiple black-box models. The results pave way for fair yet accurate sentiment detectors for multiple applications.

\end{abstract}




\maketitle

\section{Introduction}
There is an increasing concern amongst researches that machine learning models developed with the best of intentions may exhibit biases, promote inequality, or perform unfairly for unprivileged groups. With the increasing usage of such models, a considerable amount of research has been conducted to recognize and address these issues and their social impact \cite{mehrabi2019survey, sun2019mitigating}. When models show signs of bias, then these models are referred as "unfair". 

Sentiment detection is an important building block for multiple applications such as content moderation, product recommendation, misinformation detection and recently language generation \cite{park2018reducing, hutchinson2020unintended, raisi2019reduced,wolf2019transformers}. To build these models, large training corpora from varied sources are used. Unfortunately, training data might already contain some bias and that bias can transmit to the learning phase; thus unprivileged groups get unfairly impacted. While the developers of such models often assess success by measuring the accuracy factor, a few have examined the fairness aspect.  

Due to the aforementioned popularity, large companies such as Google \footnote{Google Cloud \url{https://cloud.google.com/natural-language}}, Amazon \footnote{Amazon Comprehend \url{https://aws.amazon.com/comprehend/}} and IBM \footnote{Ibm Watson \url{https://www.ibm.com/watson}} provide black-box models that can be easily incorporated into any applications to provide a sentiment score for a given text. Although sentiment detection plays a significant role in such tasks, potential discriminatory treatments might exist for different populations and since these providers are perceived as trustworthy, a severe impact could hurt large populations \cite{o2016weapons}. For example, when a model often predicts a text snippet as toxic when a female pronoun is present but fails to do so for the male pronoun, this could amplify stereotypes, disenfranchise certain groups, and yield systemic misogyny (or conversely misandry) \cite{singh2020female}. 

An important reason for success in multimedia computing is having multiple models (often involving different datasets) combined together to achieve better results. This approach has been widely used for combining multiple weak forecast models, such as in weather forecast. Similarly, for machine learning, "ensemble learning" and "multi-modal fusion" methods show astonishing results in terms of co-learning and accuracy enhancement \cite{mendes2012ensemble, atrey2010multimodal}. Recently, ensemble deep learning models have been utilized for various applications such as image, video, and speech recognition \cite{lee2017ensemble, deng2014ensemble}.

Recognizing the importance of fairness in such applications, multiple researchers have proposed various metrics and bias mitigation methods \cite{mehrabi2019survey, sun2019mitigating, nozza2019unintended, alasadi2019toward}. Many such methods either massage the incoming data  (pre-processing approaches) or change the optimization parameters within the white-box machine learning model (in-processing) \cite{calmon2017optimized}. 
As both of these are not easily possible in multiple applications (e.g., news sentiment detection using Google API), we focus on a post-processing approach for combining the results from multiple black box APIs (which we also refer to as "modalities" in this work).

In this paper, we examine the fairness aspect of three popular sentiment detection black-box models on crime news headlines in which misogynistic and/or misandristic bias might exist and we propose a method to mitigate this bias by combining the output from different models. Specifically, we examine the sentiment detection across \textbf{"gendered interaction"} in news, such as \emph{"a woman hurts a man in a bus"} vs \emph{"a man hurts a woman in a bus"}. Here, "gendered interaction" indicates that there are two actors both with clearly identified and differing gender, and there is an interaction taking place between them. In such settings, a model that produces a positive score for the first sentence but a negative score for the second is considered biased towards women (perpetrators) and unfair for men (victims) and vice versa. We apply these tests on crime news headlines dataset that has been collected specifically for this study. Experimental results show each of the publicly available APIs has inherent gender bias and also inaccuracies. On the positive side, the proposed "Flexible Fair Regression" approach was found to be useful to ameliorate both fairness and accuracy concerns.

Our main contributions in this paper are:

\begin{itemize}
\item To examine the fairness aspect of publicly available sentiment detection APIs that have been used extensively in various applications. We report that each of these models have inherent gender bias.

\item To propose an optimization method "Flexible Fair Regression" to easily allow balancing between bias and accuracy when combining the outputs from multiple (semi-accurate and semi-biased) black box models.
\item To share the newly created approach and resulting dataset for quantifying bias in "gendered interaction" scenarios
\footnote{\url{https://github.com/abdulazizasz/fairness\_sentiment}}.
\end{itemize}

Note that we consider the use of binary gender as a limitation of this work. The use of gender neutral pronouns and those inclusive of non-binary identities is still not common enough in news headlines and hence the problem of bias with binary gendered pronouns remains an important challenge \cite{badjatiya2019stereotypical, nozza2019unintended}.

The rest of the paper is organized as follows. Section 2 provides an overview of the related work in bias detection and mitigation. Then, Section 3 describes the methods including the bias measurement approach and the proposed bias mitigation strategy. The experimental setup and results are provided in Sections 4 and 5. Finally, in Section 6, a summary and future directions are shared.

\vspace{-10pt}
\section{Related Work}
There is significant research work devoted to fairness in algorithms and they can be divided into two categories: (1) bias measurement, and (2) bias mitigation.   

For the first category, a commonly used strategy is to compare the treatment differences of different privileged groups \cite{kiritchenko2018examining, rudinger2017social}. Muliple efforts use the approach of "word-swapping" where different sensitive variables are swapped in a fixed context. For example, in \cite{park2018reducing} authors use a template "\textit{I hate <identity> people}" by replacing the sensitive variable with different identities such as (gay, jewish, african, etc). This process has been used to measure bias in sentiment detection
\cite{shen2018darling, kiritchenko2018examining}, coreference resolution \cite{rudinger2018gender} and language models \cite{huang2019reducing}. This technique provides a practical approach to examine the treatment for different sentences when a sensitive variable plays an important role in forming the sentence. Another set of metrics that have been widely used for classification tasks are measuring the differences of accuracy, False Positive Rate (FPR), False Negative Rate (FNR) and so on \cite{park2018reducing, dixon2018measuring}. Unfortunately, these metrics are only applicable for classification tasks, whereas in sentiment detection the output value is typically a continuous score, therefore different statistical measures must be utilized to examine unfair treatment.

Multiple studies have examined unfair treatments in settings in which a single sensitive variable exists in a text such as ("I hate women", or "Jews are bad"). Additionally, Rudinger et al.,  \cite{rudinger2018gender} in coreference resolution examine the bias of an inferred pronoun in a text that has (two actors). However, none have studied the bias issue in sentiment detection in cases where there are two different actors involved in an interaction.   

For the second category, bias mitigation strategies can be used in different levels: pre-processing, in-processing and post-processing. Calmon et al., \cite{calmon2017optimized} propose a de-biasing method which uses a probabilistic transformation that edits the features and labels in the data with group fairness. Another pre-processing method presented by Zemel et al., \cite{zemel2013learning} focuses on learning a fair representation technique that finds a latent representation which encodes the data well but obfuscates information about sensitive attributes. Kamishima et al., \cite{kamishima2012fairness} propose an in-processing algorithm known as Prejudice Remover to decrease bias by adding a discrimination–aware regularization term to the learning objective function. Celis et al., \cite{celis2019classification} put forward the idea of a meta fair classifier that takes fairness metric as part of the input and returns a classifier optimized with respect to a fairness metric. Other efforts such as \cite{pleiss2017fairness, hardt2016equality} bring forward the idea of calibrated equalized odds which is a post-processing technique that optimizes over calibrated classifier score. 

Past work on multimodal fusion has 
proposed methods to enhance the accuracy by combining multiple modalities in data-levels, feature-levels or the decision-levels \cite{mendes2012ensemble, atrey2010multimodal}. These methods are also feasible in tasks when dealing with different black-box models that have different levels of accuracy and different level of bias. Although these approaches have shown great success in the past, they are yet to be studied with the goal of fairness enhancement.  Fairness in multimedia computing is a relatively nascent but fast-growing \cite{alasadi2019toward, singh2020legal} field and this work helps motivate and ground the need for fairness considerations in fusion research. 
The approach in this paper is inspired by multi-modal fusion proposals for co-learning with weak learners to enhance the accuracy \cite{mendes2012ensemble, atrey2010multimodal, deng2014ensemble, lee2017ensemble}. It also adapts the in-processing technique for bias mitigation using a regularizer to create a post-processing approach which can work well with multiple black box models in sentiment detection case. 


\section{Methodology}
 \subsection{Preliminaries}
We formulate the problem of fair sentiment detection as the following. We have $k$ independent black-box models and their sentiment scores $x_k \in [-1, 1] $ and a ground truth score $y \in [-1, 1]$. Besides that, let there be a sensitive variable $S$ that can divide the dataset $D = \{x_i, y_i\}^{N}_{i=1}$ into different groups, e.g., $S_{male}$, $S_{female}$. To simplify the settings, we can combine multiple $k$ column vectors modalities $\{x_{1}, \  x_{2}, \ ... x_{k} \}$ in a matrix $X$.  

In such a setting the goal of the algorithm is to minimize the loss as measured via a combination of accuracy error, bias penalty, and over-fitting penalty. 
We describe the operationalization of the different terms above in the following subsections.  



\subsection{Measuring Bias}
To measure the bias/fairness, we examine the mean difference of the scores for each black-box models with respect to a sensitive variable $S$, e.g., for a binary variable ${S^+, S^-}$ as follows:  

$$ Mean \ Difference \ (x, k) \  = \frac{\sum\limits_{x_i \in S^+} x^{k}_{i}}{| S^+ |}  - \frac{\sum\limits_{x_i \in S^-} x^{k}_{i}}{| S^- |} $$

A fair black-box model will result in zero score which means the same sentiment score has been produced regardless of the sensitive variable. Since we are focusing on the scores deviation among different groups we can use the Mean Absolute Difference/Error (MAE). Other methods proposed in the literature for measuring the bias can also be useful candidates (e.g. Correlation and delta of prediction accuracy \cite{calders2013controlling}) in other settings.

\subsection{Balancing Accuracy and Fairness}
In this project, we use linear regression to find the best parameter to combine these independent black-box models $X$ with respect to the target scores $y$. The linear regression can be formulated as an optimization problem to find the parameter $w$ that minimizes the following function:

\begin{equation}
\begin{aligned}
& \underset{w}{\textbf{minimize}}
& & MSE(w) =  \frac{1}{N} \sum_{i=1}^{N} (w * x_i - y_i)^2  \\
\end{aligned}
\end{equation}

Many variants of linear regression add a regularizer function to the regression, which keeps a check on the number of parameters being used in the modeling \cite{hoerl1970ridge}.
Further, recent efforts have proposed adding a "fairness regularizer" to the regression \cite{berk2017convex}. Here, we adapt that approach to define a regularizer which penalizes the function when the sentiment scores differ between different groups. For a binary variable, a model regularizes the  difference between $S^+$ and $S^-$. 

In other words, the modified objective function is trying to find the optimal parameter $w$ that minimizes the Mean Squared Error (MSE) along with the minimum bias between different groups. To simplify the objective function, a bias matrix $\Delta$ which contains the sentiment score difference for each modality is calculated. For instance, for a modality $k$ and a binary sensitive variable $S$, the bias vector can be calculated as follows:

$$\delta_{k} = | x_{S+}^{k} -x_{S-}^{k}  |$$

Similarly, calculating the bias for other modalities and combining them in one matrix $\Delta$ yields:  
$$
\begin{aligned}
& \Delta = 
\begin{bmatrix}
 &  & \\
\delta_{i}^{1} & . . . & \delta_{i}^{k} \\
 &  & 
\end{bmatrix}
\end{aligned}
$$

Using the $w$ and $\Delta$ the "fairness" penalty function $P$ is: 

$$
P(w) =  \frac{1}{N} \sum_{i=1}^{N} (w * \delta_i)^2
$$

Now the optimization function for "Flexible Fair Regression" is:

\begin{equation}
\begin{aligned}
& \underset{w}{\textbf{minimize}}
& & L(w) =   MSE(w) + \beta P(w) + \lambda \| w \|^2  \\
\end{aligned}
\end{equation}

where $ \| w \|^2$ is the $L_2$ norm. Hyper parameters $\lambda$ and $\beta$ are used to control over-fitting and the fairness trade-off. We call this approach "flexible fair regression" as it supports fairness in regression, and allows a system designer to flexibly pick the relative importance and thresholds that they want to assign to fairness and accuracy.

\section{Experimental Setup}
In this section, we describe the process of the data-collection, annotations, baselines, and the bias-reduction results.

\begin{figure*}[h]
  \centering
  \includegraphics[width=1.00\linewidth]{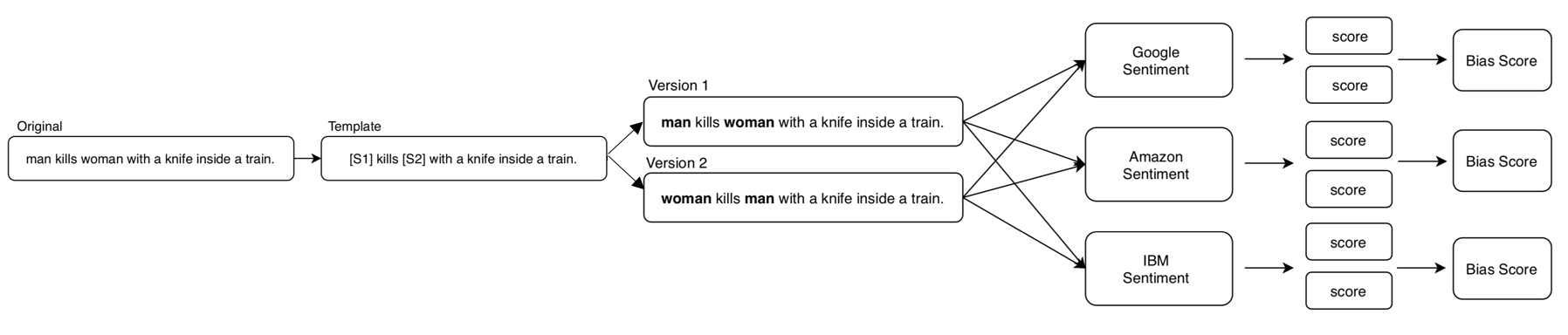}
  \caption{Constructing a template, the two versions and getting the APIs scores}
  \Description{}
\end{figure*}

\subsection{Dataset}

To construct a dataset, we collected crime news headlines from Google News API \footnote{https://news.google.com}. To do so, we used the API search criteria to retrieve only the news headlines that contain abusive verbs such as (\emph{kill, murder, slap}, etc) along with at least two different subjects (\emph{man,  woman}). Using a carefully designed list of abusive verbs \cite{wiegand2018inducing}, we collect a large number of data-points (crime news). Since in this project we are tackling the predictive learning problem as a linear regression, we need to label and assign a sentiment score to the collected dataset. Utilizing Figure-Eight \footnote{www.figure-eight.com}, we asked 10 annotators to label and score each sentence (template) and to avoid bias in the annotation process, we  anonymize the subjects in the sentence. Then, every annotator is presented with an anonymized template (see Fig. 1) and is asked to provide two pieces of information following the Valence-Arousal model \cite{russell1980circumplex}: (1) A valence label to a sentence such as "Positive or Negative" (2) An arousal score on a scale from 1 to 10. Thus, a sentence with a positive label and a score of 10 will have a sentiment score close to +1 whereas a sentence with a negative label and an arousal score of 5 will have a sentiment score -0.5. This process yielded scores in the range of [-1, 1].

Removing inconsistent annotators, and using a seed of 200 templates, we did gender swapping in the sentences where the first sentence has men as perpetrators and women as victims and vice-versa. By applying different gender identities such as ("man-woman", "male-female", etc), we used 25 of such terms provided by \cite{gonen2019lipstick}.
Thus, our corpus contains 10000 news headline sentences. The resulting dataset was then scored using Google, Amazon, and IBM APIs (see Fig. 1). Finally, we split the dataset into training and testing sets in a ratio of 70:30.

A sample of a dataset is shown in Table 1. For each black box model, we have a sentiment score for the two versions of the template along with the gender and the ground-truth score. From Table 1, we get the scores for each template by taking the average between two genders since we are aiming at finding the optimal score for each template regardless of the gender (see Table 2).

\begin{table}
  \caption{A sample from the dataset}
  \label{tab:freq}
  \begin{tabular}{cccccc}
    \toprule
    Sentence    &   $k_1$   &  $k_2$  &  $k_3$ & $S$ & $y$ \\
    \midrule
     \textbf{man hurts woman in ..} & $-0.9$ & $-0.5$ & $0.6$ & $m$ & $-0.7$\\
     \textbf{woman hurts man in ..} & $-0.7$ & $-0.8$ & $-0.9$ & $f$ &$-0.7$\\
    
  \bottomrule
\end{tabular}
\end{table}

\begin{table}
  \caption{Training dataset for each template regardless of the gender}
  \label{tab:freq}
  \begin{tabular}{ccccc}
    \toprule
    Sentence    &   $k_1$   &  $k_2$  &  $k_3$ & $y$ \\
    \midrule
     \textbf{"[S1] hurts [S2] in ..} & $-0.8$ & $-0.65$ & $0.15$ & $-0.7$\\
    
  \bottomrule
\end{tabular}
\end{table}

\subsection{Baselines}
Since the scores that are generated from these black-box models are not always accurate, a fusion process is used to increase the accuracy by combining different modalities. Related works (e.g.,  \cite{atrey2010multimodal, mendes2012ensemble}) provide various methods that are practical for fusing independent modalities. In this project we are experimenting with three such methods:

\textbf{Unweighted Average} is the basic fusion process that assumes that independent modalities are equal in terms of accuracy. Thus, the predicted sentiment score is calculated as follows:
$$
\hat{y_i} = \frac{1}{N} ( x_i^{google} + x_i^{amazon} + x_i^{ibm} ) 
$$

\textbf{Weighted Average} weights each modality based on its accuracy (for the training set): 
$$
\hat{y_i} =  w_{Google} * x_i^{google} + w_{Amazon} * x_i^{amazon} +  w_{Ibm} * x_i^{ibm} 
$$
where $\sum_{k=1}^{3} w_k = 1$. 


\textbf{Multiple Regression} is similar to our proposed method but without the "fairness" penalty term. In other word, black-box models outputs can be treated as features for another learning model.

All of the above methods are only used to optimize for the accuracy part and not considering the fairness aspect. 

\textbf{Fairness Optimization} is an additional baseline that optimizes (only) for fairness. We weight each modality by its fairness scores in the training data. The "fairness" weight for a modality $k$ is calculated as follows:

\begin{equation}
     w^{k} = \sum_{i=1}^{N}\ \mathds{1} \{ | x_{male_i}^{k} - x_{female_i}^{k} |  \leq \tau\}\
\end{equation}

Here, if the sentiment scores differ by less than $\tau=10\%$, we consider it as a fair treatment for that template. 


\section{Results And Discussion}
\subsection{Auditing Sentiment Detection APIs}
As a first step, we analyzed the results from the different black box models (sentiment detection APIs) to see if they are accurate and if there is a difference in the results obtained for sentences which are same, except for the genders represented for the perpetrators and the victims.

To evaluate accuracy (compared to ground truth labels obtained from multiple human labelers)  we use Root Mean Squared Error (RMSE). To measure bias we use Mean Absolute Error (MAE). Table 3 shows the errors in accuracy and bias levels for each individual modality. The range for sentiment scores was [-1, 1] and we see that there are noteworthy accuracy and bias issues with each modality. Further, a pairwise t-test comparing the mean sentiment scores across genders yielded statistically significant differences in all three modalities.  These issues motivate the need to combine the outputs from different modalities to improve both accuracy and fairness.

\begin{table}
  \caption{Accuracy and Fairness in the original dataset}
  \label{tab:freq}
  \begin{tabular}{ccl}
    \toprule
    API    &   Acc. Error   &    Bias     \\
    \midrule
    Google      &   $0.5611$    & $0.0590$                \\
     Amazon      &   $0.6939$    & $0.0581$                \\
      IBM      &   $0.7441$    & $0.0545$             \\
    
  \bottomrule
\end{tabular}
\end{table}

\subsection{Improving Accuracy and Fairness}

\begin{table}
  \caption{Mitigation Methods Analysis}
  \label{tab:freq}
  \begin{tabular}{ccl}
    \toprule
    Model Name    &   Acc. Error   &    Bias\\
    \midrule
    Multiple Regression          &   0.5362        &  0.0738 \\
    Unweighted Average           &   0.6302        &  0.0435 \\
    Weighted Average             &   0.6153        &  0.0447 \\
    Fairness Optimization        &   0.7051        &  0.0173 \\
   \textbf{ Our Method }         &   0.6026        &  0.0400 \\

  \bottomrule
\end{tabular}
\end{table}
We implemented the Flexible Fair Regression approach (Eq. 2) on the created dataset using Python. 
Table 4 provides the summary of the results. We can easily see the trade-off between the accuracy error and bias among these models (lower is better in both cases). Multiple Regression is performing well in terms of accuracy (Low RMSE) but has to contend with higher value for bias. Both Weighted Average and Unweighted Average methods yield higher errors in terms of accuracy than Multiple Regression but yield lower levels of bias.  
Note that it is also possible to optimize only for fairness (Fairness Optimization), and this can reduce the bias to very low value (close to zero). However, this comes with the price of a high accuracy error (see Fig. 2). 

Lastly, our method allows for flexible trade-off between accuracy and fairness (see Eq. 2).  The trade-off depends on the choice of weight parameter $\beta$ and different values of $\beta$ yield different points on the purple curve in Fig. 3. The axes for the figure are accuracy error and bias levels. Lower is better for both of them and hence, the points on the lower left corner are ideal. As shown, all the points of the purple curve (our approach with different $\beta$ values) either coincide with other baselines or strictly dominate them (i.e., yield better performance in terms of both accuracy and fairness). 

\begin{figure*}[h]
  \centering
  \includegraphics[width=1\linewidth]{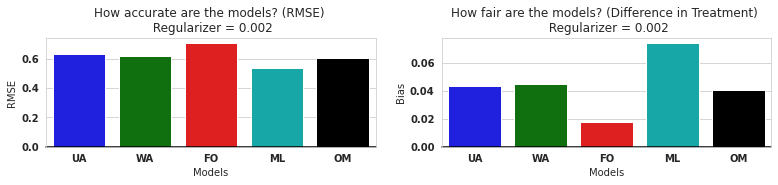}
  \caption{The accuracy error (RMSE) and the bias score (MAE) is shown for different baselines methods along with our method. The lower the bar the better. \emph{"UA: Unweighted Average, WA: Weighted Average, FO: Fairness Optimization, ML: Multiple Regression, OM: Our Method"}}
  \Description{}
\end{figure*}

\begin{figure}[h]
  \centering
  \includegraphics[width=1\linewidth]{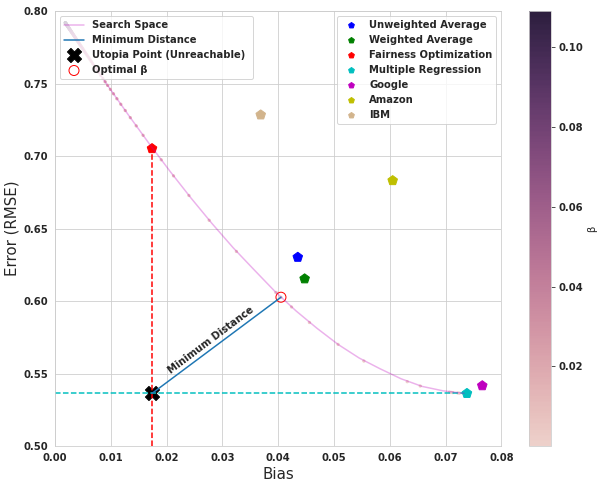}
  \caption{Using the intersected point, we choose the closest $\beta$ which is $\beta= .002$ in Our Method}
  \Description{}
\end{figure}

\begin{figure}[h]
  \centering
  \includegraphics[width=1\linewidth]{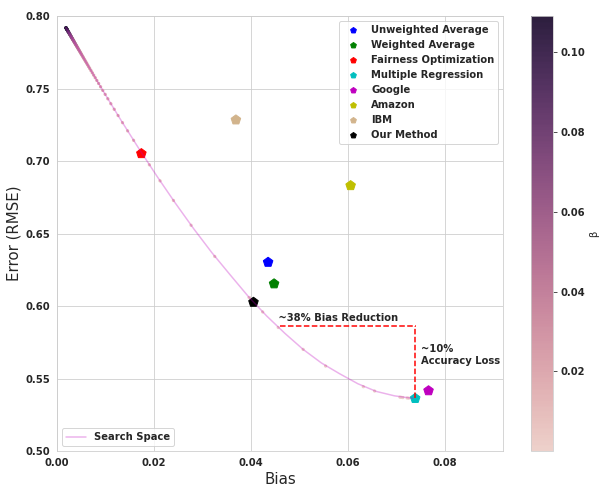}
  \caption{Trade-off between accuracy and fairness by manipulating the parameter $\tau$ (e.g., $\tau=10\%$) }
  \Description{}
\end{figure}

The Multiple Regression and Fairness Optimization approaches can be considered extreme cases of our flexible fair regression approach, such that they optimize only for accuracy or only for fairness. Any point on the purple curve (varying values of $\beta$) will provide a trade-off between these two extremes. 
To find a specific suitable candidate for $\beta$ value, we use Figure 3 to jointly consider the achievements of Fairness Optimization (FO) and Multiple Regression (MR). Hence, an ideal solution ("Utopia Point" marked as "X" in Fig. 3; unlikely to be achievable in practice), will yield accuracy error as low as MR and bias as low as FO (see Fig. 3).  
Hence, we consider the point closest (minimum distance) to the "Utopia Point" to be a suitable candidate to pick the $\beta$ value. In the current work, $\beta$ value of 0.002 is the closest point, which yields the results shown in Table 4 and Figure 2. Note that this result is pareto-optimal in the sense that there is no other feasible point that is lower in \textit{both} accuracy error and bias. This can be seen from the points in Fig. 3 (which also includes points for models which use just one modality) and Table 4.

Another possible approach to joint optimization of two factors is to budget a fixed "cost of fairness" \cite{berk2017convex}. For example, losing a portion of accuracy could lead to a gain in fairness. In Figure 4, we provide an illustration of trade-off; a ${\sim}10\%$ accuracy loss could yield a ${\sim}38\%$ reduction in bias. Hence, a practitioner with an assigned $10\%$ accuracy budget could gain up to $38\%$ in terms of fairness. Other plausible budgets and impact can also be easily computed using this approach to allow for such decision making.

\section{Conclusion}
In this work, we deploy a regularized objective function that combines independent black box models to ensure an accurate and a fair learning model for sentiment detection. Since we are dealing with disparate and independent black-box models, a fusion process helps combine multiple sources' results and build a more robust score for each template regardless of the subjects' genders. The proposed approach yields a family of pareto-optimal solutions compared to other baseline approaches. Further, our "fairness" penalty function performs well in terms of bias reduction and is more flexible compared to other baselines.  

An important limitation of this work is the focus on binary genders. Another critical challenge is that of constructing large and practical dataset, i.e., templates that cover enough amount of context in which abusive verbs occur. In this study, we only investigate how sentiment detection APIs are dealing with crime news headlines.
Additionally, the annotation process might result in biased scores; thus, to mitigate this bias, we take the average of the scores from different annotators and removed inconsistent annotators. Lastly, since we are solving an optimization function, a convexity of equation is assumed and a solver has been used to find the optimal minima. 

Despite the limitations, this paper marks a significant step forward toward fairness and accuracy in sentiment detection literature. The paper advances the fairness literature to consider multiple actor "gendered interactions", which has use cases in news analysis, abuse detection, and misinformation detection. The public dataset and the proposed flexible approach can allow for fairness in a wide variety of scenarios where semi-accurate and semi-fair black box models need to be combined to obtain fair yet accurate predictions. 

\section*{Acknowledgments}
This material is in part based upon work supported by
the National Science Foundation under Grant SES-1915790.

\bibliographystyle{ACM-Reference-Format}
\bibliography{paper-bib}

\end{document}